\documentclass[10pt,twocolumn,letterpaper]{article}

\usepackage{cvpr}              

\usepackage{graphicx}
\usepackage{amsmath}
\usepackage{amssymb}
\usepackage{booktabs}
\usepackage{multirow}
\usepackage{makecell}
\usepackage{amsmath}
\usepackage{dutchcal}
\usepackage{bbding} 
\usepackage{colortbl}
\usepackage[accsupp]{axessibility}
\usepackage{bm}
\definecolor{mygray}{gray}{.9}
\usepackage{url}

\usepackage[pagebackref,breaklinks,colorlinks]{hyperref}

\usepackage[capitalize]{cleveref}
\crefname{section}{Sec.}{Secs.}
\Crefname{section}{Section}{Sections}
\Crefname{table}{Table}{Tables}
\crefname{table}{Tab.}{Tabs.}

\def\cvprPaperID{7254} 
\def\confName{CVPR}
\def\confYear{2022}

\begin{document}

\title{Beyond Fixation: Dynamic Window Visual Transformer}

\author{Pengzhen Ren\textsuperscript{1} \quad Changlin Li\textsuperscript{2} \quad Guangrun Wang\textsuperscript{3} \quad Yun Xiao\textsuperscript{1}\thanks{Corresponding author.} \\ 
\quad Qing Du\textsuperscript{4}\footnotemark[1] \quad
Xiaodan Liang\textsuperscript{5,6} \quad Xiaojun Chang\textsuperscript{2,7}\\{\normalsize
\textsuperscript{1}Northwest University, China \quad       \textsuperscript{2}ReLER, AAII, University of Technology Sydney \quad
\textsuperscript{3}University of Oxford}\\{\normalsize  
\textsuperscript{4}South China University of Technology \quad
\textsuperscript{5}Sun Yat-sen University \quad \textsuperscript{6} PengCheng Laboratory \quad
\textsuperscript{7}RMIT University}\\
{\tt\small pzhren@foxmail.com, yxiao@nwu.edu.cn, duqing@scut.edu.cn, xiaojun.chang@uts.edu.au} \\	{\tt\small \{changlinli.ai, wanggrun, xdliang328\}@gmail.com}
}
\maketitle

\begin{abstract}
Recently, a surge of interest in visual transformers is to reduce the computational cost by limiting the calculation of self-attention to a local window. Most current work uses a fixed single-scale window for modeling by default, ignoring the impact of window size on model performance.  However, this may limit the modeling potential of these window-based models for multi-scale information. In this paper, we propose a novel method, named Dynamic Window Vision Transformer (DW-ViT). The dynamic window strategy proposed by DW-ViT goes beyond the model that employs a fixed single window setting. To the best of our knowledge, we are the first to use dynamic multi-scale windows to explore the upper limit of the effect of window settings on model performance. In DW-ViT, multi-scale information is obtained by assigning windows of different sizes to different head groups of window multi-head self-attention. Then, the information is dynamically fused by assigning different weights to the multi-scale window branches. We conducted a detailed performance evaluation on three datasets, ImageNet-1K, ADE20K, and COCO. Compared with related state-of-the-art (SoTA) methods, DW-ViT obtains the best performance. Specifically, compared with the current SoTA Swin Transformers \cite{liu2021swin}, DW-ViT has achieved consistent and substantial improvements on all three datasets with similar parameters and computational costs. In addition, DW-ViT exhibits good scalability and can be easily inserted into any window-based visual transformers.\footnote{Code release: \url{https://github.com/pzhren/DW-ViT}. This work was done when the first author interned at Dark Matter AI.}
\end{abstract}
\vspace{-1em}

\begin{figure}
\center{\includegraphics[width=0.4\textwidth] {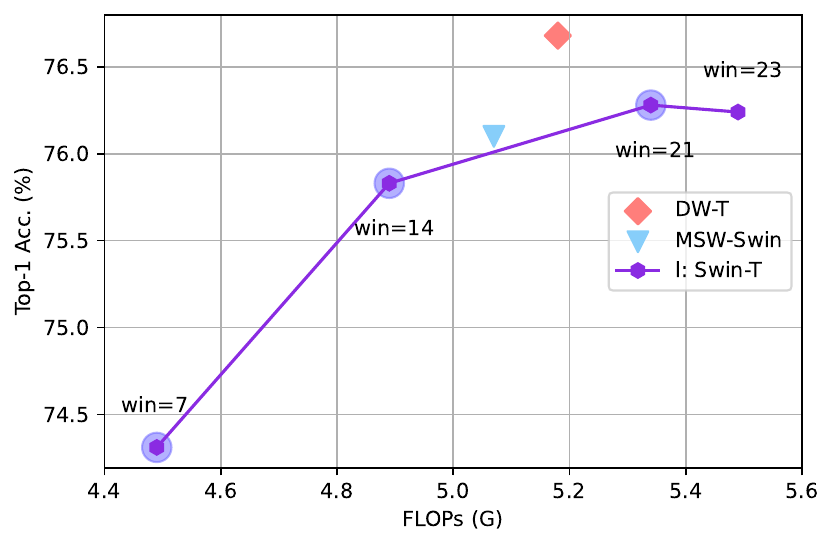}}   
\vspace{-0.8em}
   \caption{Performance comparison of DW-ViT, Swin \cite{liu2021swin} and Swin with multi-scale window (MSW-Swin) on ImageNet-1K \cite{Deng2009ImageNetAL} as the window size increases. We use a purple broken line ($l$) to indicate the performance and FLOPs changes of Swin-T \cite{liu2021swin} with a single-scale window ($win\in[7,14,21,23]$). The multi-scale windows used by MSW-Swin and DW-T are all set to $[7,14,21]$.}
 \label{fig:window_acc_flops}
 \vspace{-1.8em}
\end{figure}

\section{Introduction}
In computer vision (CV) tasks, the visual transformer represented by Vision Transformer (ViT) \cite{dosovitskiy2021an} has shown great potential. These methods have achieved impressive performance on tasks such as image classification \cite{srinivas2021bottleneck, wang2021pyramid}, semantic segmentation  \cite{wu2021p2t, liu2021polarized} and object detection \cite{liu2021swin, yang2021focal, yuan2021tokens}. 

In ViT, the complexity of the self-attention operation is proportional to the square of the number of image patches. This is unfriendly to most tasks in the CV field. Swin \cite{liu2021swin} thus proposed to limit the calculation of self-attention to a local window to reduce the computational complexity and achieved some promising results. This local window self-attention quickly attracted a significant amount of attention~\cite{lin2021cat, chu2021twins, wang2021crossformer}. 
However, most of these methods \cite{lin2021cat, chu2021twins, wang2021crossformer} use a fixed single-scale window (\eg, $win=7$) by default. The following questions accordingly arise: \textit{Is this window size optimal? Does a bigger window entail better performance? Is a multi-scale window more advantageous than a single-scale window? Furthermore, will dynamic multi-scale windows yield better results?} To answer these questions, we evaluate the impact of window sizes on the model performance. In \cref{fig:window_acc_flops}, we report the change curve ($l$) of top-1 accuracy and FLOPs (G) of Swin-T \cite{liu2021swin} under four single-scale windows ($win\in[7,14,21,23]$) on ImageNet-1K \cite{Deng2009ImageNetAL}. In Swin \cite{liu2021swin}, the window size has a very small effect on the amount of model parameters.

As shown in \cref{fig:window_acc_flops}, as the window size increases, the performance of the model is found to be significantly improved, but this is not absolutely monotonous. For example, when the window size is increased from 21 to 23, the performance of the model hardly improves or even drops. Therefore, it is not feasible to simply increase the window to improve the performance of the model. In addition, it is difficult to choose the best window size from multiple alternative window sizes. And the optimal window settings of different layers may also be different. A natural idea is to mix information from windows of different scales for prediction tasks.
Based on this idea, we design a multi-scale window multi-head self-attention (MSW-MSA) mechanism for the window-based ViT. In \cref{fig:window_acc_flops}, as shown in the results of Swin-T with MSW (MSW-Swin) and Swin-T with a single-scale window, simply introducing the MSW mechanism for the W-MSA of the transformer cannot further effectively improve the performance of the model.
For example, the performance of MSW-Swin ($win=[7,14,21]$) is lower than that of Swin-T with single-scale windows when $win=21$. It may be caused by suboptimal window settings that impairs the performance of the model. This shows that it may require more effort to protect ViT with MSW from suboptimal window settings while retaining the advantages of multi-scale windows.
On the other hand, the dynamic neural network \cite{han2021dynamic} has been favored by a large number of researchers because of its ability to adjust the structure and parameters of the model adaptively according to the input.
Moreover, the dynamic network has been successfully applied in CNN \cite{szegedy2016rethinking, szegedy2017inception, xie2017aggregated, zoph2018learning, tan2019mixconv, li2019selective} and ViT \cite{wang2021crossformer, yang2021focal, chen2021crossvit}.

\begin{figure}
\center{
    \subfloat[DW-ViT (ours)]{
    	\includegraphics[width=0.25 \textwidth]{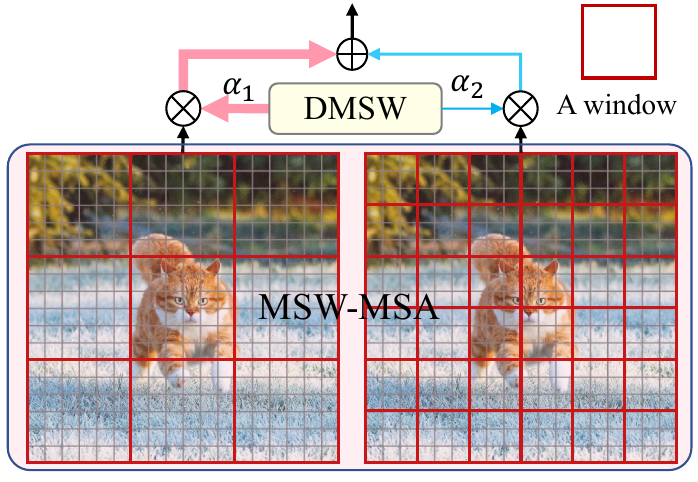}
    	\label{fig:MSWin_Swin_a_}}
    	\hspace{0.5em}
    \subfloat[Swin Transformer]{
    	\includegraphics[width=0.13 \textwidth]{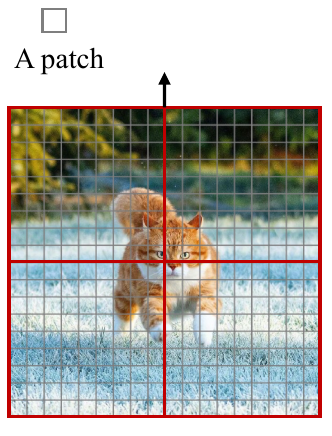}
    	\label{fig:MSWin_Swin_b_}}
   }
   \vspace{-0.7em}
   \caption{Comparison of DW-ViT's multi-scale window (\eg, $win_1=6$ and $win_2=3$) and Swin-based single-scale window (\eg, $win=9$). The number of patches in the local window is  $win \times win$. A dynamic multi-scale window (DMSW) is a dynamic adaptive window module we designed for multi-scale window multi-head self-attention (MSW-MSA). $\alpha$ is a learnable parameter of the DMSW module. $\alpha_1$ and $\alpha_2$ are a possible weight distribution scheme of DMSW.}
\label{fig:MSWin_Swin}
\vspace{-1.8em}
\end{figure}

Based on the above observations, in this paper, we propose a novel method, named Dynamic Window Vision Transformer (DW-ViT). As far as we know, it is the first method to use dynamic multi-scale windows to explore the upper limit of the impact of window settings on model performance. In DW-ViT, we first obtain multi-scale information by assigning different scale windows to different head groups of multi-head self-attention in transformer. Then, we realize the dynamic fusion of information by assigning weights to the multi-scale window branches.
In \cref{fig:MSWin_Swin}, we present a comparison of DW-ViT's multi-scale window and single-scale window approaches based on Swin \cite{liu2021swin} class methods.
More specifically, in DW-ViT, MSW-MSA is responsible for the extraction of multi-scale window information, while DMSW is responsible for the dynamic enhancement of these multi-scale information. Through the above two parts, DW-ViT can improve the model's multi-scale information modeling capabilities dynamically while ensuring relatively low computational complexity. As shown in \cref{fig:window_acc_flops}, the performance of DW-T with a dynamic window is significantly better than that of Swin-T with a single fixed-scale window, which we call "beyond fixed".
Our main contributions can be summarized as follows:
\begin{itemize}
    \item The recently popular window-based ViT mostly ignores the influence of window size on model performance. This severely limits the upper limit of the model's performance. As far as we know, we are the first to challenge this problem.
    \item We propose a novel plug-and-play module with a dynamic multi-scale window for multi-head self-attention in transformer. DW-ViT is superior to all other ViTs that use the same single-scale window and can be easily embedded into any window-based ViT.
    \item Compared with the state-of-the-art methods, DW-ViT achieves the best performance on multiple CV tasks with similar parameters and FLOPs.
\end{itemize}

\section{Related Works}

\begin{figure}
	\centering
	\includegraphics[width=0.4 \textwidth]{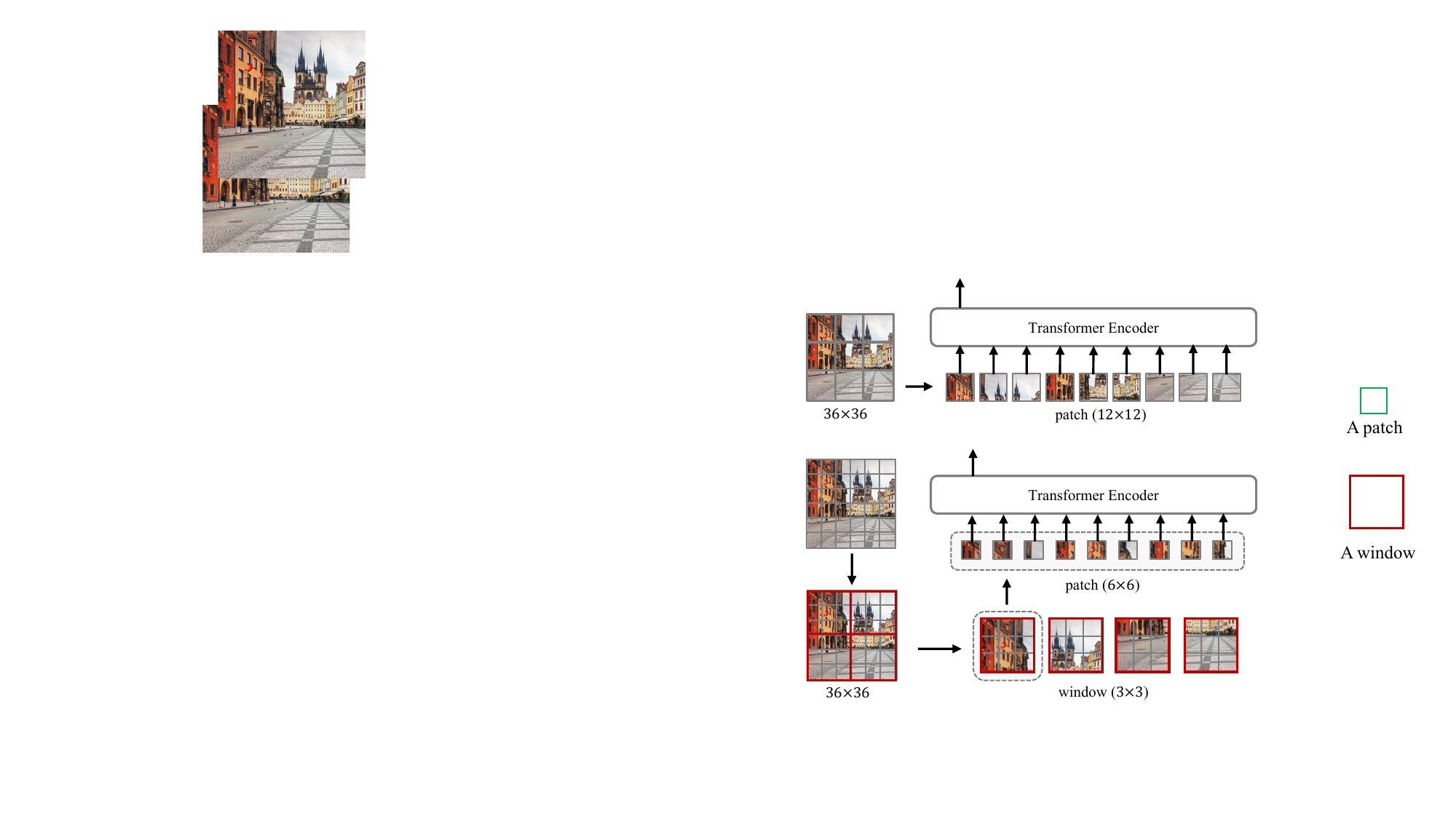}
	\vspace{-0.7em}
	\caption{In the visual transformer, a schematic diagram of the window self-attention calculation process. Assume that the number of pixels in the input image is $H\times W$ (\eg $36 \times 36$). The image is first split into $\lceil\frac{H}{p}\rceil\times \lceil\frac{W}{p}\rceil$ fixed-size patches (\eg $p=6$), and then the self-attention calculation is limited to a fixed-size window (\ie  each window has $M \times M$ patches, \eg $M=win=3$). For simplicity, patch and position embeddings are omitted here.}
	\label{fig:ViT-window}
	 \vspace{-1em}
\end{figure}

\begin{figure*}
	\centering
	\includegraphics[width=0.75 \linewidth]{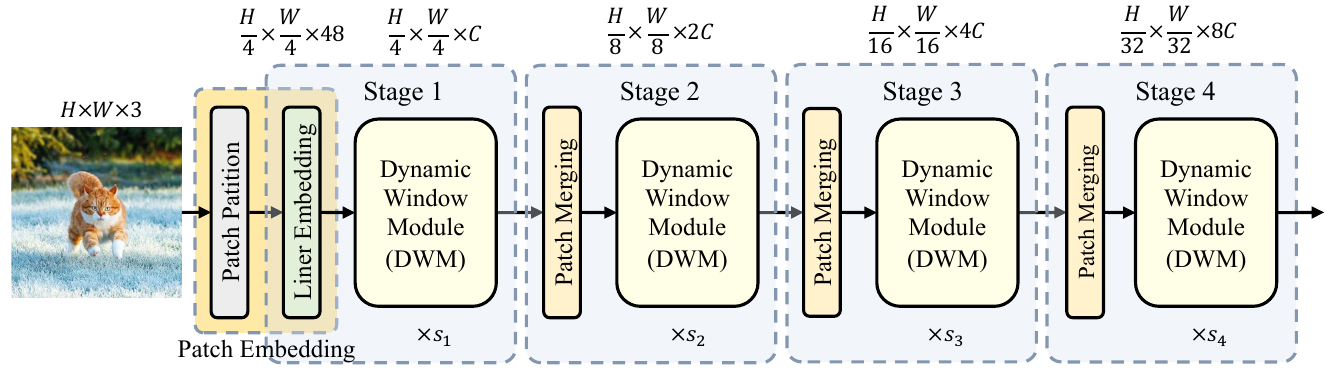}
	\vspace{-0.55em}
	\caption{The architecture of the Dynamic Window Vision Transformer (DW-ViT).}
	\label{fig:Architecture}
	\vspace{-1.8em}
\end{figure*}

\noindent\textbf{Window self-attention.} 
In the ViT context, standard self-attention splits each image into  fixed-size patches \cite{dosovitskiy2021an, wang2021pyramid, touvron2021training}. These patches are expanded as a sequence of tokens, which are then fed to the transformer encoder after being encoded. 
The calculation amount of this standard self-attention is still huge.
Subsequent work \cite{wang2021pyramid, wu2021p2t, Hu2019LocalRN} has continued to try to reduce the computational complexity of standard self-attention.
In particular, Swin \cite{liu2021swin} proposes to limit the calculation of self-attention to a local window.
This window self-attention strategy reduces the computational complexity of MSA from $\mathcal{O}{(N^2)}$ to $\mathcal{O}{(N)}$ (here $N$ is the number of image patches).
The schematic diagram of the window-based self-attention calculation process in ViT is shown in  \cref{fig:ViT-window}.
This window self-attention mechanism quickly attracted the attention of a large number of researchers \cite{chu2021twins, yang2021focal, wang2021crossformer}.
However, these works all use a fixed single-scale window. 
They ignored the impact of window size on model performance.
This may limit the upper limit of the impact of window configuration on model performance.
In \cref{fig:window_acc_flops}, the performance comparison of Swin \cite{liu2021swin} under different single-scale windows just verifies this idea.
Based on the above observations, we filled this gap and explored in detail the effect of window size on model performance, which is a supplement to the above work.

\noindent\textbf{Multi-scale information in ViT.}
Multi-scale information has been successfully applied in the field of convolution. To obtain more comprehensive information, the model not only needs small-scale information but also large-scale information. For example, Inception \cite{szegedy2015going, Szegedy2017Inceptionv4IA}, Timeception \cite{Hussein2019TimeceptionFC}, MixConv \cite{Tan2019MixConvMD} and SKNet \cite{li2019selective}, among others, obtain multi-scale information by using different sizes of convolution kernels.
In addition, some works \cite{graham2021levit, wang2021crossformer} also try to use the output of CNN as the input of ViT to improve the ability of ViT to model local information. In particular, CrossFormer \cite{wang2021crossformer} uses multi-scale convolution to provide multi-scale information for the ViT input.
Recently, due to the popularity of ViT in the CV field, many researchers have attempted to introduce multi-scale information into ViT. The pyramid structure in CNN is a widely borrowed idea. For example, T2T \cite{yuan2021tokens} reduces the length of the token sequence stage by stage by aggregating adjacent patches, while PVT \cite{wang2021pyramid} reduces the feature dimension by modifying self-attention. BossNAS \cite{li2021bossnas} searches for the downsampling position of multi-stage transformers. 
Further, P2T \cite{wu2021p2t} introduces pyramid pooling into the self-attention of the transformer. Similarly, Focal self-attention \cite{yang2021focal} also incorporates multi-scale information into the calculation of each self-attention. More directly, CrossVit \cite{chen2021crossvit} has designed a two-branch transformer encoder with image tokens of different sizes. All of these improve the model's ability to model multi-scale information to varying degrees. However, the above-mentioned method either has a large amount of calculation due to the global self-attention, or it is difficult to expand due to the complex design. In our work, we design a multi-scale window mechanism for MSA to enhance its modeling capabilities in the context of multi-scale information. This MSW-MSA strategy applies to most types of W-MSA computing and exhibits good expansion.

\noindent\textbf{Dynamic multi-branch network.}
Recently, dynamic networks \cite{han2021dynamic,li2021dynamic,li2021ds} are popular because they can flexibly adjust the structure and parameters of the network according to the input and have better adaptive capabilities.
In a dynamic multi-branch network, a common strategy is to assign corresponding weights to different branches according to their importance to achieve a large-capacity, more versatile, and flexible network structure.
For example, early works on this topic \cite{jacobs1991adaptive, Eigen2014LearningFR} used real-valued weights to dynamically rescale the representations obtained from different experts. In addition, SKNet \cite{li2019selective}, ACNet \cite{wang2019adaptively}, TreeConv \cite{wang2020grammatically}, and ResNeSt \cite{Zhang2020ResNeStSN} propose a simple split-attention mechanism that dynamically adjusts the weight of the information obtained by different convolution kernels or branches. This strategy can obtain dynamic feature representations for different samples with a small computational cost, thereby improving the model's expressive ability.
In our work, the proposed multi-scale window self-attention module has a natural affinity with the above-mentioned dynamic multi-branch network. Accordingly, we propose a dynamic multi-scale window (DMSW) module for MSW-MSA. This DMSW strategy enables DW-ViT to integrate information from windows of different scales in a dynamic manner so that the model can obtain better expressive capabilities.

\section{Method}
\subsection{Overall Architecture}
To facilitate proper comparison while maintaining its high-resolution task processing capabilities, DW-ViT follows the architectural design outlined in \cite{liu2021swin, wang2021pyramid, Zhang2021MultiScaleVL}. \cref{fig:Architecture} presents the overall architecture of DW-ViT. The model comprises four stages. To generate hierarchical feature representation, the $i$-th stage consists of a feature compression layer and $s_i$ Dynamic Window Module (DWM) transformer layers. More specifically, in Stage 1, similar to the ViT \cite{dosovitskiy2021an,liu2021swin}, the RGB image is split into non-overlapping patches (the patch size is set to $4\times 4$; that is, the compression ratio in the spatial dimension is 4). The original RGB pixel value of each patch is concatenated (\ie after patch concatenation, the dimension is $4\times 4\times 3=48$) and projected to an arbitrary dimension (denoted as $C$) through a linear embedding layer. The feature dimension of the corresponding patch embedding layer output is $\frac{H}{4}\times \frac{W}{4} \times C$. These generated patch tokens are then used as the input of the DWM transformer layers, and the number (\ie $\frac{H}{4}\times\frac{W}{4}$) of tokens remains unchanged during this process. Similarly, Stages 2–4 uses a similar structure. The difference is that the feature compression ratio of the patch merging layer in each stage is 2, while the number of channels is doubled. That is, the resolutions of the output features for Stages 2–4 are $\frac{H}{8}\times\frac{W}{8}$, $\frac{H}{16}\times\frac{W}{16}$, and $\frac{H}{32}\times\frac{W}{32}$, and the corresponding channel dimensions are $2C$, $4C$, and $8C$, respectively. The combination of output features at different stages can be used as the input of task networks such as classification, segmentation, and detection.

 \begin{figure*}
	\centering
	\includegraphics[width=0.85 \linewidth]{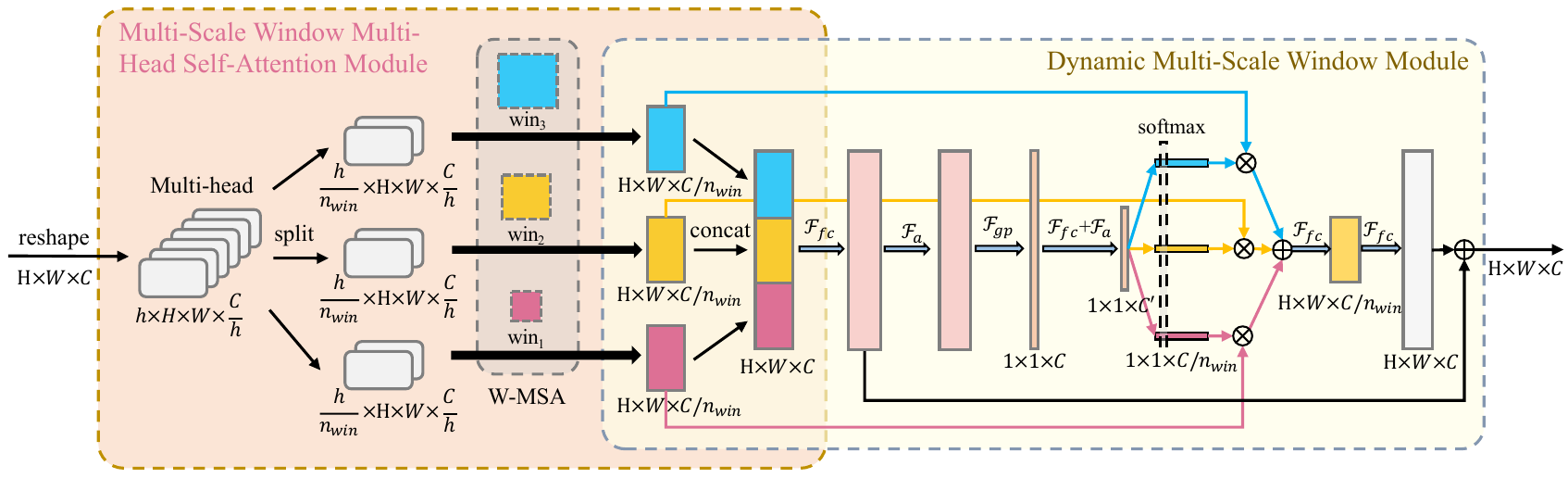}
	\vspace{-0.85 em}
	\caption{Dynamic Window Module (DWM). DWM has two main parts: Multi-Scale Window Multi-Head Self-Attention Module (MSW-MSA) and Dynamic Multi-Scale Window Module (DMSW).}
	\label{fig:OMSWin}
	\vspace{-1.5em}
\end{figure*} 

\subsection{Dynamic Window Module}

As shown in \cref{fig:OMSWin}, the DWM we designed comprises two main parts: a multi-scale window multi-head self-attention module (MSW-MSA) and a dynamic multi-scale window module (DMSW). The former is responsible for the capture of multi-scale window information, while the latter is responsible for the dynamic adaptive weighting of this information.

\subsubsection{Multi-Scale Window Multi-head Self-Attention}  
\cref{fig:OMSWin} (left) presents an architecture diagram of MSW-MSA with $h$ heads and $n_{win}$ scale windows. Here we take $h=6$ and $n_{win}=3$ as an example. The multi-head $h$ of MSA is evenly divided into $n_{win}$ groups, which perform multi-head self-attention at different scales window to capture multi-scale window information. A group of windows here can be set to $\text{Win} = \{win_{i}, i=1,...,n_{win}\}$. Specifically, assume the input feature map $\bm{x}\in \mathbb{R}^{H\times W\times C}$; we thus have the following output of MSW-MSA:
\begin{equation}
\begin{aligned}
    \bm{y}_{\text{MSW-MSA}} &=\text{MSW-MSA}(\bm{x})\\
    &= \text{Concat}(\{\text{W-MSA}_{win_i}(\hat{\bm{y}}_i)\}),\\
     \hat{\bm{y}}_i&=\text{Split}_{i}(\hat{\bm{x}})\in \mathbb{R}^{\frac{h}{n_{win}}\times H\times W\times\frac{C}{h}}, i=1,...,n_{win},\\
    \hat{\bm{x}} &= \text{Reshape}(\bm{x})\in \mathbb{R}^{h\times H\times W\times \frac{C}{h}},\\
\end{aligned}
\end{equation}
where the $i$-th branch $\hat{\bm{y}}_i$ is divided into $\lceil\frac{H}{win_i}\rceil \times \lceil\frac{W}{win_i}\rceil$ windows in the spatial dimension. Each window is expanded into a token sequence of length $win_i \times win_i$ and used as the input of the $i$-th branch W-MSA$_{win_i}$ of MSW-MSA. The structure of W-MSA is illustrated in \cref{fig:ViT-window}.  The output of W-MSA$_{win_i}$ is reconstructed as $H\times W$ in the spatial dimension, and the final output dimension is $H\times W\times \frac{C}{n_{win}}$. The outputs of these branches are concatenated in the channel dimension and used as the output of the entire MSW-MSA module.

\subsubsection{Dynamic Multi-Scale Window}

The output $\bm{y}_{\text{MSW-MSA}}\in \mathbb{R}^{H\times W\times C}$ of the multi-branch structure MSW-MSA can naturally be used as the input of DMSW. $\bm{y}_{\text{MSW-MSA}}=\text{Concat(\{W-MSA}_{win_i}(*),i=1,...,n_{win}\})$ retains the multi-scale information of window groups of different scales in the channel dimension. To this end, we designed an dynamic multi-scale window information weighting module DMSW for MSW-MSA. 

In more detail, DMSW uses the integrated information of all branches to generate corresponding weights for each branch, then integrates the information of different branches via weighting. The DMSW structure diagram is presented on the right of \cref{fig:OMSWin}. This process is divided into two main steps: \textit{Fuse} and \textit{Select}. The former is responsible for integrating the information of all branches, while the latter generates corresponding weights for each branch based on the global information and completes the fusion of branch information. 
Specifically, the details of these two parts are as follows:

\noindent\textbf{\textit{Fuse}:} 
It mainly consists of a pooling layer $F_{gp}$ and two pairs of fully connected layers $F_{fc}$ and activation layers $F_a$.
The calculation process is as follows:
\begin{equation}
\begin{aligned}
    \bm{y}_{\text{Fuse}} = &\delta_2(F_{fc_2}(F_{gp}(\delta_1( \hat{\bm{y}}_{\text{Fuse}})))),\\
    &\hat{\bm{y}}_{\text{Fuse}}=F_{fc_1}(\bm{y}_{\text{MSW-MSA}}),
\end{aligned}
\end{equation}
where $F_a=\delta$ is the GELU \cite{Hendrycks2016BridgingNA} function. The specific dimension setting is presented in \cref{fig:OMSWin} (right), where $\bm{y}_{\text{Fuse}}\in \mathbb{R}^{1\times 1\times C'}$ and $C'$ is set to $\frac{C}{2n_{win}}$.

\noindent\textbf{\textit{Select}:} It consists of two parts. The first part is composed of a set of fully connected layers $F_{\alpha}=\{F_{\alpha_i,i=1,2,...,n_{win}}\}$ and a softmax layer to generate corresponding weights for each branch, while the second contains two linear mapping layers to restore the channel dimension of the fused features. The specific calculation process is as follows:
\begin{equation}
\begin{aligned}
   &\bm{y}_{\text{Select}}=F_{fc_4}(F_{fc_3}(\sum_{i}^{n_{win}}\alpha_i\times\text{W-MSA}_{win_i}(\hat{\bm{y}}_i))),\\
   &\qquad \alpha_i=\frac{e^{F_{\alpha_i}(\bm{y}_{\text{Fuse}})}}{\sum_i^{n_{win}}e^{F_{\alpha_i}(\bm{y}_{\text{Fuse}})}},i=1,2,...,n_{win},\\
\end{aligned}
\end{equation}
where $\alpha_i\in \mathbb{R}^{1\times 1\times \frac{C}{n_{win}}}$. The DMSW module output is as follows:
\begin{equation}
    \bm{y}_{\text{DMSW}}=\bm{y}_{\text{Select}}+\hat{\bm{y}}_{\text{Fuse}}.
\end{equation}
Moreover, $\bm{y}_{\text{DMSW}}\in \mathbb{R}^{H\times W\times C}$ is also the output of the entire DWM.


\subsection{Dynamic Window Block}

The DW block is constructed by replacing the standard MSA module in the Transformer block with DWM. In addition, because DWM is designed for multi-scale information, it does not specifically design for cross-window information exchange. In the interests of simplicity, following the design presented in \cite{liu2021swin}, we retain the Swin's \cite{liu2021swin} shifted window strategy. DWM with shifted window strategy is defined as a dynamic shifted window (DSW) block. Each DWM (or DSW) block consists of two LayerNorm (LN) layers and a two-layer MLP with GELU nonlinearity. DSW achieves cross-window information exchange by moving the feature $\lfloor \frac{win}{2}\rfloor$ patches to the upper left in the spatial dimension. When the feature is reconstructed, it moves $\lfloor \frac{win}{2}\rfloor$ patches to the lower right to restore the spatial position of the feature. Alternate stacking of DWM and DSW is used to avoid a decline in information exchange. Specifically, 
two successive DWM blocks are calculated as follows:
\begin{equation}
\begin{aligned}
    &\hat{\bm{z}}^l=\text{DWM}(\text{LN}(\bm{z}^{l-1}))+\bm{z}^{l-1},\\
    &\bm{z}^l=\text{MLP(LN}(\hat{\bm{z}}^l))+\hat{\bm{z}}^l,\\
    &\hat{\bm{z}}^{l+1}=\text{DSW(LN}(\bm{z}^{l}))+\bm{z}^{l},\\
    &\bm{z}^{l+1}=\text{MLP(LN}(\hat{\bm{z}}^{l+1}))+\hat{\bm{z}}^{l+1},\\
\end{aligned}
\end{equation}
where $\hat{\bm{z}}^l$ and $\bm{z}^l$ respectively define the output of the DWM (DSW) module and MLP module in the $l$-th block.

\begin{table*}
    \centering
    \scriptsize
    \setlength{\tabcolsep}{8pt}
    \begin{tabular}{c|c|c|c|c}
    \toprule
        &Output Size&Layer Name & DW-T & DW-B\\
        \hline
        \multirow{3}{*}{Stage 1} & \multirow{3}{*}{$\frac{H}{4}\times \frac{W}{4}$} & Patch Embedding & $p_1=4;C_1=96$ & $p_1=4;C_1=128$ \\
        \cline{3-5}
        && DWM & $\left[\begin{array}{c}
             \text{Win}_1= [7,14,21] \\
             h_1=3,C_1=96
        \end{array}\right] \times 2$ & $\left[\begin{array}{c}
             \text{Win}_1= [7,12,17,22]\\
             h_1=4,C_1=128
        \end{array}\right] \times 2$ \\
        \hline
        
        \multirow{3}{*}{Stage 2} & \multirow{3}{*}{$\frac{H}{8}\times \frac{W}{8}$} & Patch Merging & $p_2=2;C_2=192 $ & $p_2=2;C_2=256 $ \\
        \cline{3-5}
        && DWM & $\left[\begin{array}{c}
             \text{Win}_2= [7,14,21]\\
             h_2=6,C_2=192
        \end{array}\right] \times 2$ & $\left[\begin{array}{c}
             \text{Win}_2= [7,12,17,22]\\
             h_2=8,C_2=256
        \end{array}\right] \times 2$ \\
        \hline
        
        \multirow{3}{*}{Stage 3} & \multirow{3}{*}{$\frac{H}{16}\times \frac{W}{16}$} & Patch Merging & $p_3=2;C_3=384 $ & $p_3=2;C_3=512 $ \\
        \cline{3-5}
        && DWM & $\left[\begin{array}{c}
             \text{Win}_3= [7,14,14]\\
             h_3=12,C_3=384
        \end{array}\right] \times 6$ & $\left[\begin{array}{c}
             \text{Win}_3= [7,12,14,14]\\
             h_3=16,C_3=512
        \end{array}\right] \times 18$ \\
        \hline
        
        \multirow{3}{*}{Stage 4} & \multirow{3}{*}{$\frac{H}{32}\times \frac{W}{32}$} & Patch Merging & $p_4=2;C_4=768 $ & $p_4=2;C_4=1024 $ \\
        \cline{3-5}
        && DWM & $\left[\begin{array}{c}
             \text{Win}_4= [7,7,7]\\
             h_4=24,C_4=768
        \end{array}\right] \times 2$ & $\left[\begin{array}{c}
             \text{Win}_4= [7,7,7,7]\\
             h_4=32,C_4=1024
        \end{array}\right] \times 2$ \\
    \bottomrule
    \end{tabular}
    \vspace{-0.7em}
    \caption{Configuration details of DW-ViT. Here, $p_i\times p_i$ is the size of the patch in the $i$-th stage, and is also the downsampling ratio of the feature in the spatial dimension. $C_i$ is the number of feature channels, while $\text{Win}_i$ and $h_i$ are the window combination used by the MSW-MSA module and the number of heads used by the MSA in transformer respectively.}
    \vspace{-1.8em}
    \label{tab:OMSWin_setting}
\end{table*}

\noindent\textbf{Position encoding.} For a local window with $M\times M$ patches, following \cite{Raffel2020ExploringTL, Bao2020UniLMv2PL,liu2021swin}, we added a set of relative position bias $B=\{B_i\in \mathbb{R}^{M_i^2\times M_i^2},i=1,2,...,n_{win}\}$ to the similarity calculation of each head of DWM self-attention. For the W-MSA$_{win_i}$ of the $i$-th scale local window, we have the window self-attention calculation of $Q_i$ as follows:
\begin{equation}
    \text{Attention}(Q_i,K_i,V_i)=\text{SoftMax}(\frac{Q_iK_i^T}{\sqrt{d}}+B_i)V_i,
\end{equation}
where $Q_i,K_i,V_i\in \mathbb{R}^{M_i^2\times d}$ are \textit{query}, \textit{key}, and \textit{value} matrices, while $M_i^2$ is the number of patches in the $i$-th scale window, and $d$ is the $Q_i/K_i$ dimension. In addition, we parameterized a bias matrix set $\hat{B}=\{\hat{B}_i,i=1,...,n_{win}\}$. Specifically, for $\hat{B_i}$, because the relative position on each axis lies in the range of $[-M_i+1,M_i-1]$, a small-sized bias matrix $\hat{B_i}\in \mathbb{R}^{(2M_i-1)\times (2M_i-1)}$ is parameterized, and the values in $B_i$ are taken from $\hat{B_i}$.

\subsection{Model Configuration}

To facilitate fair comparison, following \cite{liu2021swin}, we set the two configuration models as DW-T and DW-B. Their configuration details are summarized in \cref{tab:OMSWin_setting}. In particular, according to the results in \cref{fig:window_acc_flops} and the size of the output features in each stage on ImageNet \cite{Deng2009ImageNetAL}, for the DW-T with three heads in the first stage, we set $\text{Win}_1=[7,14,21]$. For Stages 2–4, we adjust the window according to the size of the output feature of each stage (when the size of the window and the output feature are equal, the standard self-attention is calculated at this time). Similarly, for DW-B, $\text{Win}_1=[7,12,17,22]$. For all experiments, the query dimension of each head is $d=32$, while the expansion layer of each MLP is $\alpha=4$.

\subsection{Complexity Analysis}

The computational complexity of the DWM block is composed of two main parts: $\Omega(\text{SMW-MSA})$ and $\Omega(\text{DMSW})$. For an image with $\mathcal{h}\times \mathcal{w}$ patches, their computational complexity is as follows\footnote{The calculation of SoftMax is ignored here.}:
\begin{equation}
  \Omega(\text{SMW-MSA})=4\mathcal{hw}C^2+2\mathcal{hw}\frac{C}{n_{win}}\sum_i^{n_{win}}win_i^2. 
\end{equation}
\begin{equation}
  \Omega(\text{DMSW})=(1+\mathcal{hw}(1+\frac{1}{n_{win}}))\frac{C^2}{n_{win}}. 
\end{equation}
The total computational complexity of DWM is as follows:
\begin{equation}
\begin{aligned}
     \Omega(\text{DWM})=&\Omega(\text{SMW-MSA})+\Omega(\text{DMSW}) \\
     =&(1+4n_{win}+\frac{\mathcal{hw}+n_{win}}{\mathcal{hw}n_{win}})\frac{\mathcal{hw}}{n_{win}}C^2+\\
     &2\mathcal{hw}\frac{C}{n_{win}}\sum_i^{n_{win}}win_i^2.
\end{aligned}
\end{equation}
Since both $win_i$ and $n_{win}$ are constants, the total computational complexity of DWM does not significantly increase.
The computational complexity of DWM is still $\mathcal{O}(N)$.

\section{Experiments}
We conduct a performance comparison with the state-of-the-art (SoTA) methods on an upstream task, ImageNet-1K image classification \cite{Deng2009ImageNetAL}, and two downstream tasks: semantic segmentation on ADE20K \cite{Zhou2018SemanticUO}, and object detection and instance segmentation on COCO 2017 \cite{Lin2014MicrosoftCC}. Finally, we ablate the important modules of DW-ViT.

\subsection{Image Classification on ImageNet-1K}

\noindent\textbf{Experimental Settings}
We benchmark DW-ViT on ImageNet-1K \cite{Deng2009ImageNetAL}. ImageNet-1K contains 1.28M training images and 50K test images from 1000 categories. To test the effectiveness of DW-ViT and conduct a fair comparison with similar methods \cite{liu2021swin,chen2021crossvit,chu2021twins}, we carefully avoid using any tricks that provide unfair advantage \cite{Touvron2021GoingDW, Jiang2021TokenLT}. Specifically, following the settings in \cite{liu2021swin, chu2021twins}, DW-ViT was trained for 300 epochs with a batch size of 1024 using the AdamW optimizer \cite{Loshchilov2019DecoupledWD}. The cosine decay learning rate scheduler and 20 epochs of a linear warm-up are used. The initial learning rate and weight decay are set to 0.001 and 0.05, respectively. In training, \cite{Touvron2021TrainingDI}'s augmentation and regularization strategies are used. Following the settings in \cite{liu2021swin}, the repeated enhancement \cite{Hoffer2020AugmentYB} and EMA \cite{Polyak1992AccelerationOS} strategy are abandoned.

\begin{table}
    \centering
    \scriptsize
    \begin{tabular}{l|c|c|c}
    \toprule
    \multirow{2}{*}{Method}  & \multirow{2}{*}{\makecell[c]{\#param. (M)}} & \multirow{2}{*}{\makecell[c]{FLOPs (G)}}  & \multirow{2}{*}{\makecell{\makecell{Top-1 (\%)}}} \\
    &&&\\\hline
    \multicolumn{4}{c}{ConvNet} \\\hline
    ResNet50\cite{he2016deep}&26&4.1&76.6\\
    ResNet101\cite{he2016deep}&45&7.9&78.2\\\hline
    X50-32x4d\cite{xie2017aggregated}&25&4.3&77.9\\
    X101-32x4d\cite{xie2017aggregated}&44&8.0&78.7\\\hline
    RegNetY-4G \cite{radosavovic2020designing} & 21 & 4.0 & 80.0\\
    RegNetY-8G \cite{radosavovic2020designing}  & 39 & 8.0 & 81.7\\
    RegNetY-16G \cite{radosavovic2020designing}  & 84 & 16 & 82.9\\\hline
    \multicolumn{4}{c}{Transformer} \\\hline
    DeiT-Small/16 \cite{touvron2021training} & 22 & 4.6 & 79.9\\
    CrossViT-S \cite{chen2021crossvit} & 27 & 5.6 & 81.0\\
    T2T-ViT-14 \cite{yuan2021tokens} & 22 & 5.2 & 81.5\\
    TNT-S \cite{han2021transformer} & 24 & 5.2 & 81.3\\
    CoaT Mini \cite{xu2021co} & 10 & 6.8 & 80.8\\
    PVT-Small \cite{wang2021pyramid} & 25 & 3.8 & 79.8\\
    CPVT-GAP \cite{yuan2021tokens} & 23 & 4.6 & 81.5\\
    CrossFormer-S$^\dagger$ \cite{wang2021crossformer} & 28 & 4.5 & 81.5\\
    Swin-T \cite{liu2021swin} & 28 & 4.5 & 81.3\\
    \cellcolor{mygray}DW-T & \cellcolor{mygray}30 & \cellcolor{mygray}5.2 & \cellcolor{mygray}\textbf{82.0}\\\hline

    ViT-Base/16 \cite{dosovitskiy2020image} &87&17.6&77.9\\
    DeiT-Base/16 \cite{touvron2021training} & 87&17.6&81.8\\
    T2T-ViT-24 \cite{yuan2021tokens}& 64 & 14.1 & 82.3\\
    CrossViT-B \cite{chen2021crossvit}&105&21.2&82.2\\
    TNT-B \cite{han2021transformer} & 66 & 14.1 & 82.8\\
    CPVT-B \cite{chu2021conditional}& 88 &17.6 &82.3 \\
    PVT-Large \cite{wang2021pyramid} & 61 & 9.8 & 81.7\\ 
    Swin-B \cite{liu2021swin} & 88 & 15.4 & 83.3\\
    \rowcolor{mygray} DW-B & 91 & 17.0 & \textbf{83.8}\\
    \bottomrule
    \end{tabular}
    \vspace{-0.7em}
    \caption{Performance comparison on ImageNet-1K. All models are trained and evaluated at $224\times 224$ resolution. $\text{CrossFormer-S}^\dagger$ shows the performance in the case of single-scale embedding.}
    \vspace{-2.5em}
    \label{tab:ImageNet}
\end{table}

\noindent\textbf{Results}
\cref{tab:ImageNet} reports the performance comparison of DW-ViT and state-of-the-art methods on ImageNet-1K. Methods of comparison include the classic and the latest ConvNet-based \cite{he2016deep, xie2017aggregated, radosavovic2020designing} and Transformer-based \cite{liu2021swin, wang2021crossformer, chen2021crossvit} models. All models are trained and evaluated at $224\times 224$ resolution. 
As shown in \cref{tab:ImageNet}, with similar parameters and FLOPs, DW-ViT still has obvious advantages compared with other current state-of-the-art methods.
Specifically, compared with Transformer baseline DeiT \cite{touvron2021training}, the performance of DW-T and DW-B are improved by 2.1\% and 2.0\%, respectively.
At the same time, under the same settings, compared with Swin \cite{liu2021swin}, DW-T and DW-B also achieved performance gains of 0.7 and 0.5 points, respectively, with the help of dynamic windows. 
This shows that DW-ViT as a general visual feature extractor can obtain better feature representation.
In addition, it is worth mentioning that as an independent module, DWM can be flexibly embedded in any window-based ViT model \cite{wang2021crossformer, chu2021twins, lin2021cat} like Swin \cite{liu2021swin} to improve the model's dynamic modeling capabilities for multi-scale information.
Compared with these ViTs \cite{wang2021crossformer, chu2021twins, lin2021cat} that use a fixed single-scale window, DWM enables DW-ViT to have a larger model capacity and perform better in terms of adaptability and scalability.

\subsection{Semantic Segmentation on ADE20K}
\vspace{-0.5em}
\begin{table}
\scriptsize
\resizebox{\linewidth}{!}{
\begin{tabular}{l|l|cc|cc}
\toprule
\multirow{2}{*}{Backbone}    & \multirow{2}{*}{Method}  & \multirow{2}{*}{\makecell[c]{\#param.\\(M)}} & \multirow{2}{*}{\makecell{FLOPs\\(G)}} & \multirow{2}{*}{\makecell{mIoU}} & \multirow{2}{*}{\makecell{+MS}} \\
&&&&&\\\hline

ResNet-101 \cite{he2016deep} & DANet \cite{nam2017dual}  & 69     & 1119  & 45.3 & -   \\
ResNet-101 & OCRNet \cite{yuan2020object}  & 56     & 923  & 44.1 & -   \\
ResNet-101 & DLab.v3+ \cite{chen2018encoder}  & 63     & 1021  & 44.1 & -   \\
ResNet-101 &ACNet\cite{fu2019adaptive}   & -       &    -  & 45.9 & -    \\
ResNet-101 & DNL\cite{yin2020disentangled} & 69    & 1249  & 46.0 & -   \\
ResNet-101 &UperNet \cite{xiao2018unified}   & 86      &  1029 & 44.9 & -    \\\hline
HRNet-w48 \cite{sun2019deep} & DLab.v3+ \cite{chen2018encoder} &71 & 664 & 45.7\\
ResNeSt-101\cite{Zhang2020ResNeStSN}&DLab.v3+\cite{chen2018encoder}& 66 & 1051 & 46.9 &-\\
ResNeSt-200\cite{Zhang2020ResNeStSN}&DLab.v3+\cite{chen2018encoder}&88&1381&48.4&-\\\hline
PVT-S \cite{wang2021pyramid} & S-FPN \cite{kirillov2019panoptic} & 28 & - & 39.8 & \\
PVT-M & S-FPN & 48 & 219 & 41.6 & - \\
PVT-L & S-FPN & 65 & 283 & 42.1 & - \\
CAT-S \cite{lin2021cat} & S-FPN & 41 & 214 & 42.8 & - \\
CAT-B & S-FPN & 55 & 276 & 44.9 & -\\
Swin-T\cite{liu2021swin}&UperNet\cite{xiao2018unified}&60&945&44.5&45.8\\
Swin-B\cite{liu2021swin}&UperNet\cite{xiao2018unified}&121&1188&48.1&49.7\\\hline

\rowcolor{mygray}DW-T&UperNet\cite{xiao2018unified}&61&953&45.7&46.9\\
\rowcolor{mygray}DW-B&UperNet\cite{xiao2018unified}&125&1200&\textbf{48.7}&\textbf{50.3}\\
\bottomrule
\end{tabular}
}
\vspace{-0.7em}
\caption{Performance comparison on the ADE20K \cite{Zhou2018SemanticUO} val. The single-scale and multi-scale evaluation results are presented in the last two columns. The FLOPs (G) are calculated at an input resolution of $1024\times 1024$.}
\vspace{-2.5em}
\label{tab:ADE20K}
\end{table}

ADE20K \cite{Zhou2018SemanticUO} is also a widely used semantic segmentation dataset. It contains 20K training images, 2K verification images, and 3K test images, covering a total of 150 semantic categories. DW-ViT and UperNet \cite{xiao2018unified} in mmsegmentation \cite{mmseg2020} are used as the backbone and segmentation methods respectively. The pre-trained backbone used is DW-ViT trained on ImageNet-1K. Following the settings in \cite{liu2021swin}, the input size of the image is $512\times 512$, AdamW \cite{Loshchilov2019DecoupledWD} is used as the optimizer (the initial learning rate is $6\times 10^{-5}$, weight decay is 0.01, and a linear learning rate decay is used), and the model is trained with a batch size of 16 and 160K iterations. For multi-scale evaluation (+MS), the scaling ratio is between 0.5 and 1.75.

The performance comparison between DW-ViT and other methods on ADE20K val is shown in \cref{tab:ADE20K}. 
As shown in \cref{tab:ADE20K}, DW-ViT achieves the best performance compared to many state-of-the-art methods.
Specifically, under similar FLOPs and parameters, compared with Swin \cite{liu2021swin}, DW-ViT improves the single-scale evaluation by 1.2 and 0.6 points, respectively. Compared with other methods, DW-ViT has also obtained competitive results. Compared with Swin, DW-ViT has a more obvious advantage (\eg $0.7 \rightarrow 1.2$) in ADE20K than in ImageNet. This shows that the dynamic window mechanism of DW-ViT has more obvious advantages in downstream tasks such as more complex image datasets.

\subsection{Object Detection on COCO}

\begin{table*}
    \centering
    \scriptsize
    \setlength{\tabcolsep}{10pt}
    \begin{tabular}{l|c|c|ccc|ccc}
    \toprule
    Method&\makecell[c]{\#param. (M)} & \makecell[c]{FLOPs (G)} & AP$^{\text{box}}$&AP$_{50}^{\text{box}}$&AP$_{75}^{\text{box}}$&AP$^{\text{mask}}$&AP$_{50}^{\text{mask}}$&AP$_{75}^{\text{mask}}$\\\hline
    \multicolumn{9}{c}{Mask R-CNN \cite{he2017mask}}
    \\\hline
    ResNet50 \cite{he2016deep}& 44 & 260 & \makecell{41.0} &61.7 & 44.9 & 37.1 & 58.4 & 40.1 \\
    PVT-Small \cite{wang2021pyramid} & 44 & 245 & 43.0 & 65.3 & 46.9 & 39.9 & 62.5 & 42.8\\
    ViL-Small \cite{Zhang2021MultiScaleVL} & 45 & 174 & 43.4 & 64.9 & 47.0 & 39.6 & 62.1 & 42.4\\
    Swin-T \cite{liu2021swin}& 48 & 264 & 46.0 & 68.2 & 50.2 & 41.6 & 65.1 & 44.8\\
    \rowcolor{mygray} DW-T & 49 & 275 &\textbf{46.7}&\textbf{69.1}&\textbf{51.4} &\textbf{42.4}& \textbf{66.2} & \textbf{45.6}\\\hline
    ResNeXt101-64x4d \cite{xie2017aggregated} & 102 & 493 &44.4 & 64.9 & 48.8 & 39.7 & 61.9 & 42.6\\
    PVT-Large \cite{wang2021pyramid} & 81 & 364 & 44.5 & 66.0 & 48.3 & 40.7 & 63.4 & 43.7\\
    ViL-Base \cite{Zhang2021MultiScaleVL} & 76.1 & 365 & 45.7 & 67.2 & 49.9 & 41.3 & 64.4 & 44.5 \\
    Swin-Base \cite{liu2021swin} & 107 & 496 & 48.5 & 69.8 & 53.2 & 43.4 & 66.8 & 46.9\\
    \rowcolor{mygray} DW-B & 111 & 505 & \textbf{49.2} & \textbf{70.6} & \textbf{54.0} & \textbf{44.0} & \textbf{68.0} & \textbf{47.7}\\
    \bottomrule
    \multicolumn{9}{c}{Cascade Mask R-CNN \cite{cai2018cascade,he2017mask} } \\\hline
    DeiT-S$^\dagger$\cite{touvron2021training}&80&889 &\makecell{48.0}&\makecell{67.2}&\makecell{51.7} & 41.4 & 64.2 & 44.3\\
    ResNet50\cite{he2016deep} & 82 & 739 & \makecell{46.3}&\makecell{64.3}&\makecell{50.5}& 40.1 & 61.7 & 43.4\\
    Swin-T\cite{liu2021swin} & 86 & 745 &\makecell{50.5}&\makecell{69.3}&\makecell{54.9}& 43.7 & 66.6 & 47.1  \\
    \rowcolor{mygray} DW-T & 87 & 754 & \textbf{51.5} & \textbf{70.5} & \textbf{55.9}  & \textbf{44.7} & \textbf{67.8} & \textbf{48.5}\\\hline
    X101-64 \cite{xie2017aggregated} & 140 & 972 &\makecell{48.3}&\makecell{66.4}&\makecell{52.3} & 41.7 & 64.0 & 45.1 \\
    Swin-B \cite{liu2021swin} & 145 & 982
    &\makecell{51.9}&\makecell{70.9}&\makecell{56.5} & 45.0 & 68.4 & 48.7\\
    \rowcolor{mygray} DW-B& 149 & 992 & \textbf{52.9} & \textbf{71.6} & \textbf{57.5} & \textbf{45.7} & \textbf{69.0} & \textbf{50.0}\\
    \bottomrule
    \end{tabular}
    \vspace{-0.7em}
    \caption{Performance comparison of object detection and instance segmentation on the COCO2017 val dataset. Two object detection frameworks are used: Mask R-CNN \cite{he2017mask} and Cascade Mask R-CNN \cite{cai2018cascade}. The FLOPs (G) are calculated at an input resolution of $ 1280\times 800 $. $^\dagger$ indicates that additional deconvolution layers are used to generate hierarchical features.}
    \vspace{-1.8em}
    \label{tab:COCO}
\end{table*}

Further, we benchmark DW-ViT on object detection and instance segmentation with COCO 2017 \cite{Lin2014MicrosoftCC}. COCO contains 118K training, 5K validation, and 20K test images. The pre-trained model used is DW-ViT trained on ImageNet-1K. DW-ViT is used as the visual backbone and is then plugged into a representative object detection framework. We here consider two representative object detection frameworks: Mask R-CNN \cite{he2017mask} and Cascade Mask R-CNN \cite{cai2018cascade}. All models are trained on the training images and the results are reported on the validation set. The same settings were used for all frameworks. Specifically, we use multi-scale training \cite{carion2020end, sun2021sparse}, the AdamW \cite{Loshchilov2019DecoupledWD} optimizer (the initial learning rate, weight decay and batch size are 0.0001, 0.05, and 16), and a 3 $\times$ schedule (it has 36 epochs, and the learning rate decays by 10 $\times$ between epochs 27 and 33). It is implemented based on MMDetection \cite{chen2019mmdetection}. 

The performance comparison of object detection and instance segmentation on the COCO2017 val dataset is shown in \cref{tab:COCO}. 
Compared with other state-of-the-art methods, DW-ViT achieves the best performance in both object detection frameworks.
Specifically, compared with the Transformer baseline DeiT-S \cite{touvron2021training}, DW-T is improved by 3.5 points.
Compared with Swin \cite{liu2021swin}, DW-ViT has achieved an improvement of more than 0.7 points in object detection and instance segmentation under the two object detection frameworks. At the same time, compared with Swin, the parameters and FOLPs of DW-ViT have not increased significantly, which once again demonstrates the superiority of the dynamic window mechanism.
In addition, the results of the two detection frameworks show that DW-ViT can be easily embedded into different frameworks like other backbones.

\subsection{Ablation Study}
\vspace{-0.5em}
\begin{table}
    \centering
    \scriptsize
    \setlength{\tabcolsep}{3pt}
    \begin{tabular}{l|c|c|c|c}
    \toprule
    Method & Window & \makecell{\#param. (M) }&\makecell{FLOPs (G)} & \makecell{Top-1 (\%)}\\\hline
    Swin-T&\makecell{7\\11\\14\\17\\21\\23} &\makecell{28.29\\28.31\\28.34\\28.35\\28.36\\28.36} & \makecell{4.49\\4.69\\4.89\\5.06\\5.34\\5.49} & \makecell{74.31\\75.18\\75.83\\76.31\\76.28\\76.24}\\\midrule
    DW-T & DMSW\\\hline
    \makecell{MSW-MSA\\($[7,14,21]$)} & \makecell{1\\-\\\checkmark}&\makecell{29.05\\28.33\\29.77} & \makecell{5.18\\5.07\\5.18} & \makecell{73.43\\76.10\\\textbf{76.68}}\\
    \bottomrule
    \end{tabular}
    \vspace{-0.7em}
    \caption{Performance comparison of Swin and DW-ViT on ImageNet-1K \cite{Deng2009ImageNetAL} under different window and module settings.}
    \label{tab:window_acc}
    \vspace{-1.8em}
\end{table}

To explore the effects of each component of DW-ViT, we compared the performance of Swin-T with single-scale window, MSW-Swin, and DW-ViT with and without DMSW mechanism. Specifically, we set $epoch=50$; for all other settings, we adopt the default settings presented Swin \cite{liu2021swin}.
Single-scale windows are taken from $[7,11,14,17,21,23]$, and multi-scale windows are set to $[7,14,21]$\footnote{We adopted the original settings in Swin \cite{liu2021swin} and modified only the window size. When the window size is larger than the input feature, the global self-attention is performed at this time.}. Their performance on ImageNet-1K \cite{Deng2009ImageNetAL} are shown in \cref{tab:window_acc}. 

In \cref{tab:window_acc}, DMSW shows three states ('1', '-', '\checkmark'). MSW-MSA + '1' refers to removing the dynamic weight generation and directly assigning the same weight ($\frac{1}{3}$) to all branches. MSW-MSA + '-' (MSW-Swin) denotes removing the entire DMSW module, while, MSW-MSA + '\checkmark' means normal DW-T. The performance of MSW-Swin is lower than that of Swin-T with $win=21$. This may be due to the sub-optimal window setting that impairs the performance of the model to a certain extent. The performance comparison between DW-T and MSW-MSA + '1' further shows that this dynamic window mechanism achieves a very significant improvement (\ie 3.3\%). 
In addition, with the help of the dynamic window mechanism, the performance of DW-ViT is better than all ViTs that use the same single-scale window. This shows that this dynamic window weighting mechanism does play a very important role in DW-ViT.

\section{Conclusion}
\vspace{-0.5em}
The size of the window has an important impact on the performance of the model. There is currently very little systematic study of window size in the window-based ViT works. 
In this paper, we challenged this problem for the first time.
Based on our insightful observations on the above issues, we propose a novel dynamic multi-scale window mechanism for W-MSA to obtain the optimal window configuration, thereby enhancing the model's dynamic modeling capabilities for multi-scale information.
With the help of the dynamic window mechanism, the performance of DW-ViT is found to be better than all ViTs that use the same single-scale window, with the proposed approach achieving good results on multiple CV tasks. 
At the same time, DWM has good scalability, and can thus be easily inserted into any window-based ViT as a module.
\vspace{-0.5em}
\section{Discussion}
\vspace{-0.5em}
\noindent\textbf{Potential negative societal impact:}
As a general visual feature extractor, DW-ViT has shown good performance on multiple CV tasks. However, due to the domain gap between different tasks, when the model is transferred to other tasks, some fine adjustments may still be needed.

\noindent\textbf{Limitation:}
These are a few issues that we need to improve in the future: 
(1) Although DW-ViT has shown good performance on multiple vision tasks. But compared with the single-scale window self-attention mechanism \cite{liu2021swin}, DWM still introduces a small number of additional parameters and calculations. (2) In addition, as far as DWM's dynamic window mechanism is concerned, part of the computational budget is still allocated to suboptimal optional windows. However, an ideal strategy is to allocate the entire computational budget to the most potential windows at each layer of the network.

\vspace{-0.5em}
\section*{Acknowledgment}
\vspace{-0.5em}
This work was partially supported  in part by National Key R\&D Program of China under Grant No.2020AAA0109700, NSFC under Grant (No.61972315 and No.61976233), Guangdong Province Basic and Applied Basic Research (Regional Joint Fund-Key) Grant No.2019B1515120039, Guangdong Outstanding Youth Fund (Grant No. 2021B1515020061), Australian Research Council (ARC) Discovery Early Career Researcher Award (DECRA) under DE190100626, Shaanxi Province International Science and Technology Cooperation Program Project-Key Projects No.2022KWZ-14, Ministry of Science and Technology Foundation Project 2020AAA0106900 and Key Realm R\&D Program of Guangzhou 202007030007 and Open Fund from Alibaba.

{\small
\bibliographystyle{ieee_fullname}
\bibliography{Bib.bib}

\begin{thebibliography}{10}\itemsep=-1pt

\bibitem{Bao2020UniLMv2PL}
Hangbo Bao, Li Dong, Furu Wei, Wenhui Wang, Nan Yang, Xiaodong Liu, Yu Wang,
  Songhao Piao, Jianfeng Gao, Ming Zhou, and Hsiao-Wuen Hon.
\newblock Unilmv2: Pseudo-masked language models for unified language model
  pre-training.
\newblock In {\em ICML}, 2020.

\bibitem{cai2018cascade}
Zhaowei Cai and Nuno Vasconcelos.
\newblock Cascade r-cnn: Delving into high quality object detection.
\newblock In {\em Proceedings of the IEEE conference on computer vision and
  pattern recognition}, pages 6154--6162, 2018.

\bibitem{carion2020end}
Nicolas Carion, Francisco Massa, Gabriel Synnaeve, Nicolas Usunier, Alexander
  Kirillov, and Sergey Zagoruyko.
\newblock End-to-end object detection with transformers.
\newblock In {\em European Conference on Computer Vision}, pages 213--229.
  Springer, 2020.

\bibitem{chen2021crossvit}
Chun-Fu Chen, Quanfu Fan, and Rameswar Panda.
\newblock Crossvit: Cross-attention multi-scale vision transformer for image
  classification.
\newblock {\em arXiv preprint arXiv:2103.14899}, 2021.

\bibitem{chen2019mmdetection}
Kai Chen, Jiaqi Wang, Jiangmiao Pang, Yuhang Cao, Yu Xiong, Xiaoxiao Li,
  Shuyang Sun, Wansen Feng, Ziwei Liu, Jiarui Xu, et~al.
\newblock Mmdetection: Open mmlab detection toolbox and benchmark.
\newblock {\em arXiv preprint arXiv:1906.07155}, 2019.

\bibitem{chen2018encoder}
Liang-Chieh Chen, Yukun Zhu, George Papandreou, Florian Schroff, and Hartwig
  Adam.
\newblock Encoder-decoder with atrous separable convolution for semantic image
  segmentation.
\newblock In {\em Proceedings of the European conference on computer vision
  (ECCV)}, pages 801--818, 2018.

\bibitem{chu2021twins}
Xiangxiang Chu, Zhi Tian, Yuqing Wang, Bo Zhang, Haibing Ren, Xiaolin Wei,
  Huaxia Xia, and Chunhua Shen.
\newblock Twins: Revisiting the design of spatial attention in vision
  transformers.
\newblock {\em arXiv preprint arXiv:2104.13840}, 1(2):3, 2021.

\bibitem{chu2021conditional}
Xiangxiang Chu, Zhi Tian, Bo Zhang, Xinlong Wang, Xiaolin Wei, Huaxia Xia, and
  Chunhua Shen.
\newblock Conditional positional encodings for vision transformers.
\newblock {\em arXiv preprint arXiv:2102.10882}, 2021.

\bibitem{mmseg2020}
MMSegmentation Contributors.
\newblock {MMSegmentation}: Openmmlab semantic segmentation toolbox and
  benchmark.
\newblock \url{https://github.com/open-mmlab/mmsegmentation}, 2020.

\bibitem{Deng2009ImageNetAL}
Jia Deng, Wei Dong, Richard Socher, Li-Jia Li, K. Li, and Li Fei-Fei.
\newblock Imagenet: A large-scale hierarchical image database.
\newblock In {\em CVPR}, 2009.

\bibitem{dosovitskiy2020image}
Alexey Dosovitskiy, Lucas Beyer, Alexander Kolesnikov, Dirk Weissenborn,
  Xiaohua Zhai, Thomas Unterthiner, Mostafa Dehghani, Matthias Minderer, Georg
  Heigold, Sylvain Gelly, et~al.
\newblock An image is worth 16x16 words: Transformers for image recognition at
  scale.
\newblock {\em arXiv preprint arXiv:2010.11929}, 2020.

\bibitem{dosovitskiy2021an}
Alexey Dosovitskiy, Lucas Beyer, Alexander Kolesnikov, Dirk Weissenborn,
  Xiaohua Zhai, Thomas Unterthiner, Mostafa Dehghani, Matthias Minderer, Georg
  Heigold, Sylvain Gelly, Jakob Uszkoreit, and Neil Houlsby.
\newblock An image is worth 16x16 words: Transformers for image recognition at
  scale.
\newblock In {\em International Conference on Learning Representations}, 2021.

\bibitem{Eigen2014LearningFR}
David Eigen, Marc'Aurelio Ranzato, and Ilya Sutskever.
\newblock Learning factored representations in a deep mixture of experts.
\newblock {\em CoRR}, abs/1312.4314, 2014.

\bibitem{fu2019adaptive}
Jun Fu, Jing Liu, Yuhang Wang, Yong Li, Yongjun Bao, Jinhui Tang, and Hanqing
  Lu.
\newblock Adaptive context network for scene parsing.
\newblock In {\em Proceedings of the IEEE/CVF International Conference on
  Computer Vision}, pages 6748--6757, 2019.

\bibitem{graham2021levit}
Ben Graham, Alaaeldin El-Nouby, Hugo Touvron, Pierre Stock, Armand Joulin,
  Herv{\'e} J{\'e}gou, and Matthijs Douze.
\newblock Levit: a vision transformer in convnet's clothing for faster
  inference.
\newblock {\em arXiv preprint arXiv:2104.01136}, 2021.

\bibitem{han2021transformer}
Kai Han, An Xiao, Enhua Wu, Jianyuan Guo, Chunjing Xu, and Yunhe Wang.
\newblock Transformer in transformer.
\newblock {\em arXiv preprint arXiv:2103.00112}, 2021.

\bibitem{han2021dynamic}
Yizeng Han, Gao Huang, Shiji Song, Le Yang, Honghui Wang, and Yulin Wang.
\newblock Dynamic neural networks: A survey.
\newblock {\em arXiv preprint arXiv:2102.04906}, 2021.

\bibitem{he2017mask}
Kaiming He, Georgia Gkioxari, Piotr Doll{\'a}r, and Ross Girshick.
\newblock Mask r-cnn.
\newblock In {\em Proceedings of the IEEE international conference on computer
  vision}, pages 2961--2969, 2017.

\bibitem{he2016deep}
Kaiming He, Xiangyu Zhang, Shaoqing Ren, and Jian Sun.
\newblock Deep residual learning for image recognition.
\newblock In {\em Proceedings of the IEEE conference on computer vision and
  pattern recognition}, pages 770--778, 2016.

\bibitem{Hendrycks2016BridgingNA}
Dan Hendrycks and Kevin Gimpel.
\newblock Bridging nonlinearities and stochastic regularizers with gaussian
  error linear units.
\newblock {\em ArXiv}, abs/1606.08415, 2016.

\bibitem{Hoffer2020AugmentYB}
Elad Hoffer, Tal Ben-Nun, Itay Hubara, Niv Giladi, Torsten Hoefler, and Daniel
  Soudry.
\newblock Augment your batch: Improving generalization through instance
  repetition.
\newblock {\em 2020 IEEE/CVF Conference on Computer Vision and Pattern
  Recognition (CVPR)}, pages 8126--8135, 2020.

\bibitem{Hu2019LocalRN}
Han Hu, Zheng Zhang, Zhenda Xie, and Stephen Ching-Feng Lin.
\newblock Local relation networks for image recognition.
\newblock {\em 2019 IEEE/CVF International Conference on Computer Vision
  (ICCV)}, pages 3463--3472, 2019.

\bibitem{Hussein2019TimeceptionFC}
Noureldien Hussein, Efstratios Gavves, and Arnold W.~M. Smeulders.
\newblock Timeception for complex action recognition.
\newblock {\em 2019 IEEE/CVF Conference on Computer Vision and Pattern
  Recognition (CVPR)}, pages 254--263, 2019.

\bibitem{jacobs1991adaptive}
Robert~A Jacobs, Michael~I Jordan, Steven~J Nowlan, and Geoffrey~E Hinton.
\newblock Adaptive mixtures of local experts.
\newblock {\em Neural computation}, 3(1):79--87, 1991.

\bibitem{Jiang2021TokenLT}
Zihang Jiang, Qibin Hou, Li Yuan, Daquan Zhou, Xiaojie Jin, Anran Wang, and
  Jiashi Feng.
\newblock Token labeling: Training a 85.4\% top-1 accuracy vision transformer
  with 56m parameters on imagenet.
\newblock {\em ArXiv}, abs/2104.10858, 2021.

\bibitem{kirillov2019panoptic}
Alexander Kirillov, Ross Girshick, Kaiming He, and Piotr Doll{\'a}r.
\newblock Panoptic feature pyramid networks.
\newblock In {\em Proceedings of the IEEE/CVF Conference on Computer Vision and
  Pattern Recognition}, pages 6399--6408, 2019.

\bibitem{li2021bossnas}
Changlin Li, Tao Tang, Guangrun Wang, Jiefeng Peng, Bing Wang, Xiaodan Liang,
  and Xiaojun Chang.
\newblock Bossnas: Exploring hybrid cnn-transformers with block-wisely
  self-supervised neural architecture search.
\newblock In {\em Proceedings of the IEEE/CVF International Conference on
  Computer Vision}, pages 12281--12291, 2021.

\bibitem{li2021ds}
Changlin Li, Guangrun Wang, Bing Wang, Xiaodan Liang, Zhihui Li, and Xiaojun
  Chang.
\newblock Ds-net++: Dynamic weight slicing for efficient inference in cnns and
  transformers.
\newblock {\em arXiv preprint arXiv:2109.10060}, 2021.

\bibitem{li2021dynamic}
Changlin Li, Guangrun Wang, Bing Wang, Xiaodan Liang, Zhihui Li, and Xiaojun
  Chang.
\newblock Dynamic slimmable network.
\newblock In {\em Proceedings of the IEEE/CVF Conference on Computer Vision and
  Pattern Recognition}, pages 8607--8617, 2021.

\bibitem{li2019selective}
Xiang Li, Wenhai Wang, Xiaolin Hu, and Jian Yang.
\newblock Selective kernel networks.
\newblock In {\em Proceedings of the IEEE/CVF Conference on Computer Vision and
  Pattern Recognition}, pages 510--519, 2019.

\bibitem{lin2021cat}
Hezheng Lin, Xing Cheng, Xiangyu Wu, Fan Yang, Dong Shen, Zhongyuan Wang, Qing
  Song, and Wei Yuan.
\newblock Cat: Cross attention in vision transformer.
\newblock {\em arXiv preprint arXiv:2106.05786}, 2021.

\bibitem{Lin2014MicrosoftCC}
Tsung-Yi Lin, Michael Maire, Serge~J. Belongie, James Hays, Pietro Perona, Deva
  Ramanan, Piotr Doll{\'a}r, and C.~Lawrence Zitnick.
\newblock Microsoft coco: Common objects in context.
\newblock In {\em ECCV}, 2014.

\bibitem{liu2021polarized}
Huajun Liu, Fuqiang Liu, Xinyi Fan, and Dong Huang.
\newblock Polarized self-attention: Towards high-quality pixel-wise regression.
\newblock {\em arXiv preprint arXiv:2107.00782}, 2021.

\bibitem{liu2021swin}
Ze Liu, Yutong Lin, Yue Cao, Han Hu, Yixuan Wei, Zheng Zhang, Stephen Lin, and
  Baining Guo.
\newblock Swin transformer: Hierarchical vision transformer using shifted
  windows.
\newblock {\em arXiv preprint arXiv:2103.14030}, 2021.

\bibitem{Loshchilov2019DecoupledWD}
Ilya Loshchilov and Frank Hutter.
\newblock Decoupled weight decay regularization.
\newblock In {\em ICLR}, 2019.

\bibitem{nam2017dual}
Hyeonseob Nam, Jung-Woo Ha, and Jeonghee Kim.
\newblock Dual attention networks for multimodal reasoning and matching.
\newblock In {\em Proceedings of the IEEE conference on computer vision and
  pattern recognition}, pages 299--307, 2017.

\bibitem{Polyak1992AccelerationOS}
Boris~T. Polyak and Anatoli~B. Juditsky.
\newblock Acceleration of stochastic approximation by averaging.
\newblock {\em Siam Journal on Control and Optimization}, 30:838--855, 1992.

\bibitem{radosavovic2020designing}
Ilija Radosavovic, Raj~Prateek Kosaraju, Ross Girshick, Kaiming He, and Piotr
  Doll{\'a}r.
\newblock Designing network design spaces.
\newblock In {\em Proceedings of the IEEE/CVF Conference on Computer Vision and
  Pattern Recognition}, pages 10428--10436, 2020.

\bibitem{Raffel2020ExploringTL}
Colin Raffel, Noam~M. Shazeer, Adam Roberts, Katherine Lee, Sharan Narang,
  Michael Matena, Yanqi Zhou, Wei Li, and Peter~J. Liu.
\newblock Exploring the limits of transfer learning with a unified text-to-text
  transformer.
\newblock {\em ArXiv}, abs/1910.10683, 2020.

\bibitem{srinivas2021bottleneck}
Aravind Srinivas, Tsung-Yi Lin, Niki Parmar, Jonathon Shlens, Pieter Abbeel,
  and Ashish Vaswani.
\newblock Bottleneck transformers for visual recognition.
\newblock In {\em Proceedings of the IEEE/CVF Conference on Computer Vision and
  Pattern Recognition}, pages 16519--16529, 2021.

\bibitem{sun2019deep}
Ke Sun, Bin Xiao, Dong Liu, and Jingdong Wang.
\newblock Deep high-resolution representation learning for human pose
  estimation.
\newblock In {\em Proceedings of the IEEE/CVF Conference on Computer Vision and
  Pattern Recognition}, pages 5693--5703, 2019.

\bibitem{sun2021sparse}
Peize Sun, Rufeng Zhang, Yi Jiang, Tao Kong, Chenfeng Xu, Wei Zhan, Masayoshi
  Tomizuka, Lei Li, Zehuan Yuan, Changhu Wang, et~al.
\newblock Sparse r-cnn: End-to-end object detection with learnable proposals.
\newblock In {\em Proceedings of the IEEE/CVF Conference on Computer Vision and
  Pattern Recognition}, pages 14454--14463, 2021.

\bibitem{szegedy2017inception}
Christian Szegedy, Sergey Ioffe, Vincent Vanhoucke, and Alexander~A Alemi.
\newblock Inception-v4, inception-resnet and the impact of residual connections
  on learning.
\newblock In {\em Thirty-first AAAI conference on artificial intelligence},
  2017.

\bibitem{Szegedy2017Inceptionv4IA}
Christian Szegedy, Sergey Ioffe, Vincent Vanhoucke, and Alexander~Amir Alemi.
\newblock Inception-v4, inception-resnet and the impact of residual connections
  on learning.
\newblock In {\em AAAI}, 2017.

\bibitem{szegedy2015going}
Christian Szegedy, Wei Liu, Yangqing Jia, Pierre Sermanet, Scott Reed, Dragomir
  Anguelov, Dumitru Erhan, Vincent Vanhoucke, and Andrew Rabinovich.
\newblock Going deeper with convolutions.
\newblock In {\em Proceedings of the IEEE conference on computer vision and
  pattern recognition}, pages 1--9, 2015.

\bibitem{szegedy2016rethinking}
Christian Szegedy, Vincent Vanhoucke, Sergey Ioffe, Jon Shlens, and Zbigniew
  Wojna.
\newblock Rethinking the inception architecture for computer vision.
\newblock In {\em Proceedings of the IEEE conference on computer vision and
  pattern recognition}, pages 2818--2826, 2016.

\bibitem{tan2019mixconv}
Mingxing Tan and Quoc~V Le.
\newblock Mixconv: Mixed depthwise convolutional kernels.
\newblock {\em arXiv preprint arXiv:1907.09595}, 2019.

\bibitem{Tan2019MixConvMD}
Mingxing Tan and Quoc~V. Le.
\newblock Mixconv: Mixed depthwise convolutional kernels.
\newblock {\em ArXiv}, abs/1907.09595, 2019.

\bibitem{touvron2021training}
Hugo Touvron, Matthieu Cord, Matthijs Douze, Francisco Massa, Alexandre
  Sablayrolles, and Herv{\'e} J{\'e}gou.
\newblock Training data-efficient image transformers \& distillation through
  attention.
\newblock In {\em International Conference on Machine Learning}, pages
  10347--10357. PMLR, 2021.

\bibitem{Touvron2021TrainingDI}
Hugo Touvron, Matthieu Cord, Matthijs Douze, Francisco Massa, Alexandre
  Sablayrolles, and Herv'e J'egou.
\newblock Training data-efficient image transformers \& distillation through
  attention.
\newblock In {\em ICML}, 2021.

\bibitem{Touvron2021GoingDW}
Hugo Touvron, Matthieu Cord, Alexandre Sablayrolles, Gabriel Synnaeve, and
  Herv'e J'egou.
\newblock Going deeper with image transformers.
\newblock {\em ArXiv}, abs/2103.17239, 2021.

\bibitem{wang2020grammatically}
Guangrun Wang, Guangcong Wang, Keze Wang, Xiaodan Liang, and Liang Lin.
\newblock Grammatically recognizing images with tree convolution.
\newblock In {\em Proceedings of the 26th ACM SIGKDD International Conference
  on Knowledge Discovery \& Data Mining}, pages 903--912, 2020.

\bibitem{wang2019adaptively}
Guangrun Wang, Keze Wang, and Liang Lin.
\newblock Adaptively connected neural networks.
\newblock In {\em Proceedings of the IEEE/CVF Conference on Computer Vision and
  Pattern Recognition}, pages 1781--1790, 2019.

\bibitem{wang2021pyramid}
Wenhai Wang, Enze Xie, Xiang Li, Deng-Ping Fan, Kaitao Song, Ding Liang, Tong
  Lu, Ping Luo, and Ling Shao.
\newblock Pyramid vision transformer: A versatile backbone for dense prediction
  without convolutions.
\newblock {\em arXiv preprint arXiv:2102.12122}, 2021.

\bibitem{wang2021crossformer}
Wenxiao Wang, Lu Yao, Long Chen, Deng Cai, Xiaofei He, and Wei Liu.
\newblock Crossformer: A versatile vision transformer based on cross-scale
  attention.
\newblock {\em arXiv preprint arXiv:2108.00154}, 2021.

\bibitem{wu2021p2t}
Yu-Huan Wu, Yun Liu, Xin Zhan, and Ming-Ming Cheng.
\newblock P2t: Pyramid pooling transformer for scene understanding.
\newblock {\em arXiv preprint arXiv:2106.12011}, 2021.

\bibitem{xiao2018unified}
Tete Xiao, Yingcheng Liu, Bolei Zhou, Yuning Jiang, and Jian Sun.
\newblock Unified perceptual parsing for scene understanding.
\newblock In {\em Proceedings of the European Conference on Computer Vision
  (ECCV)}, pages 418--434, 2018.

\bibitem{xie2017aggregated}
Saining Xie, Ross Girshick, Piotr Doll{\'a}r, Zhuowen Tu, and Kaiming He.
\newblock Aggregated residual transformations for deep neural networks.
\newblock In {\em Proceedings of the IEEE conference on computer vision and
  pattern recognition}, pages 1492--1500, 2017.

\bibitem{xu2021co}
Weijian Xu, Yifan Xu, Tyler Chang, and Zhuowen Tu.
\newblock Co-scale conv-attentional image transformers.
\newblock {\em arXiv preprint arXiv:2104.06399}, 2021.

\bibitem{yang2021focal}
Jianwei Yang, Chunyuan Li, Pengchuan Zhang, Xiyang Dai, Bin Xiao, Lu Yuan, and
  Jianfeng Gao.
\newblock Focal self-attention for local-global interactions in vision
  transformers.
\newblock {\em arXiv preprint arXiv:2107.00641}, 2021.

\bibitem{yin2020disentangled}
Minghao Yin, Zhuliang Yao, Yue Cao, Xiu Li, Zheng Zhang, Stephen Lin, and Han
  Hu.
\newblock Disentangled non-local neural networks.
\newblock In {\em European Conference on Computer Vision}, pages 191--207.
  Springer, 2020.

\bibitem{yuan2021tokens}
Li Yuan, Yunpeng Chen, Tao Wang, Weihao Yu, Yujun Shi, Zihang Jiang, Francis~EH
  Tay, Jiashi Feng, and Shuicheng Yan.
\newblock Tokens-to-token vit: Training vision transformers from scratch on
  imagenet.
\newblock {\em arXiv preprint arXiv:2101.11986}, 2021.

\bibitem{yuan2020object}
Yuhui Yuan, Xilin Chen, and Jingdong Wang.
\newblock Object-contextual representations for semantic segmentation.
\newblock In {\em Computer Vision--ECCV 2020: 16th European Conference,
  Glasgow, UK, August 23--28, 2020, Proceedings, Part VI 16}, pages 173--190.
  Springer, 2020.

\bibitem{Zhang2020ResNeStSN}
Hang Zhang, Chongruo Wu, Zhongyue Zhang, Yi Zhu, Zhi-Li Zhang, Haibin Lin, Yue
  Sun, Tong He, Jonas Mueller, R. Manmatha, Mu Li, and Alex Smola.
\newblock Resnest: Split-attention networks.
\newblock {\em ArXiv}, abs/2004.08955, 2020.

\bibitem{Zhang2021MultiScaleVL}
Pengchuan Zhang, Xiyang Dai, Jianwei Yang, Bin Xiao, Lu Yuan, Lei Zhang, and
  Jianfeng Gao.
\newblock Multi-scale vision longformer: A new vision transformer for
  high-resolution image encoding.
\newblock {\em ArXiv}, abs/2103.15358, 2021.

\bibitem{Zhou2018SemanticUO}
Bolei Zhou, Hang Zhao, Xavier Puig, Sanja Fidler, Adela Barriuso, and Antonio
  Torralba.
\newblock Semantic understanding of scenes through the ade20k dataset.
\newblock {\em International Journal of Computer Vision}, 127:302--321, 2018.

\bibitem{zoph2018learning}
Barret Zoph, Vijay Vasudevan, Jonathon Shlens, and Quoc~V Le.
\newblock Learning transferable architectures for scalable image recognition.
\newblock In {\em Proceedings of the IEEE conference on computer vision and
  pattern recognition}, pages 8697--8710, 2018.

\end{thebibliography}
}

\end{document}